\documentclass[conference,a4paper]{IEEEtran}
\IEEEoverridecommandlockouts
\usepackage{cite}
\usepackage{amsmath,amssymb,amsfonts}
\usepackage{algorithmic}
\usepackage{graphicx}
\usepackage{textcomp}
\usepackage{xcolor}
\usepackage{svg}
\usepackage{here}
\usepackage{nidanfloat}
\usepackage[]{hyperref}
\usepackage[hang,small,bf]{caption}
\usepackage[subrefformat=parens]{subcaption}
\def\BibTeX{{\rm B\kern-.05em{\sc i\kern-.025em b}\kern-.08em
    T\kern-.1667em\lower.7ex\hbox{E}\kern-.125emX}}
\begin{document}

\title{Prediction of Seismic Intensity Distributions 

Using Neural Networks\\}

\author{\IEEEauthorblockN{\IEEEauthorblockA{Koyu Mizutani\textsuperscript{\dag *}, Haruki Mitarai\textsuperscript{\ddag *}, Kakeru Miyazaki\textsuperscript{\ddag *}, Ryugo Shimamura\textsuperscript{\ddag}, Soichiro Kumano\textsuperscript{\S}, and Toshihiko Yamasaki\textsuperscript{\S}}
}


\IEEEauthorblockA{\textit{
        \dag\ddag Faculty of Engineering, 
        \S Graduate School of Information Science and Technology
    } \\
    \textit{
        The University of Tokyo},         Tokyo, Japan\\
        \dag\S \{mizutani, kumano, yamasaki\}@cvm.t.u-tokyo.ac.jp\\
        \ddag \{mitarai-haruki520, kakekakemiya920, rshimamura169\}@g.ecc.u-tokyo.ac.jp\\
    \scalebox{0.8}{* These authors contributed equally.}
    }
}

\maketitle

\begin{abstract}
    The ground motion prediction equation is commonly used to predict the seismic intensity distribution. 
    However, it is not easy to apply this method to seismic distributions affected by underground plate structures, which are commonly known as abnormal seismic distributions.
    This study proposes a hybrid of regression and classification approaches using neural networks. The proposed model treats the distributions as 2-dimensional data like an image. Our method can accurately predict seismic intensity distributions, even abnormal distributions.
\end{abstract}


\section{Introduction}
    Predicting the seismic intensity distributions of earthquakes is important for evaluating the risk of seismic hazards. The ground motion prediction equations~(GMPEs) is a standard tool for assessing seismic ground-motion intensity\cite{DOUGLAS2016203, morikawa2013new}.
    Some studies have applied machine learning models to the prediction of the ground-motion intensity\cite{derras2012adapting, Hybrid}.
    However, these approaches are unsuitable for predicting the seismic intensity distributions affected by regional underground structures, such as abnormal seismic distributions.
    
    In this study, we propose a new data-driven method that treats epicenter data and seismic intensity distributions as 2-dimensional data, similar to an image. 
    Our model is a hybrid of the regression and classification models. 
    We compared the performance of the proposed model with only classification or only regression models in terms of the following metrics: correlation coefficient, mean squared error~(MSE), and F-score. The proposed model achieved well-balanced values for all three metrics: correlation coefficient of 0.78, MSE of 0.39, and F-score of 0.61. Furthermore, our method accurately predicts seismic intensity distributions, even abnormal distributions.

\section{Methods}
    \begin{figure}[tb]
        \includegraphics[width = \linewidth]{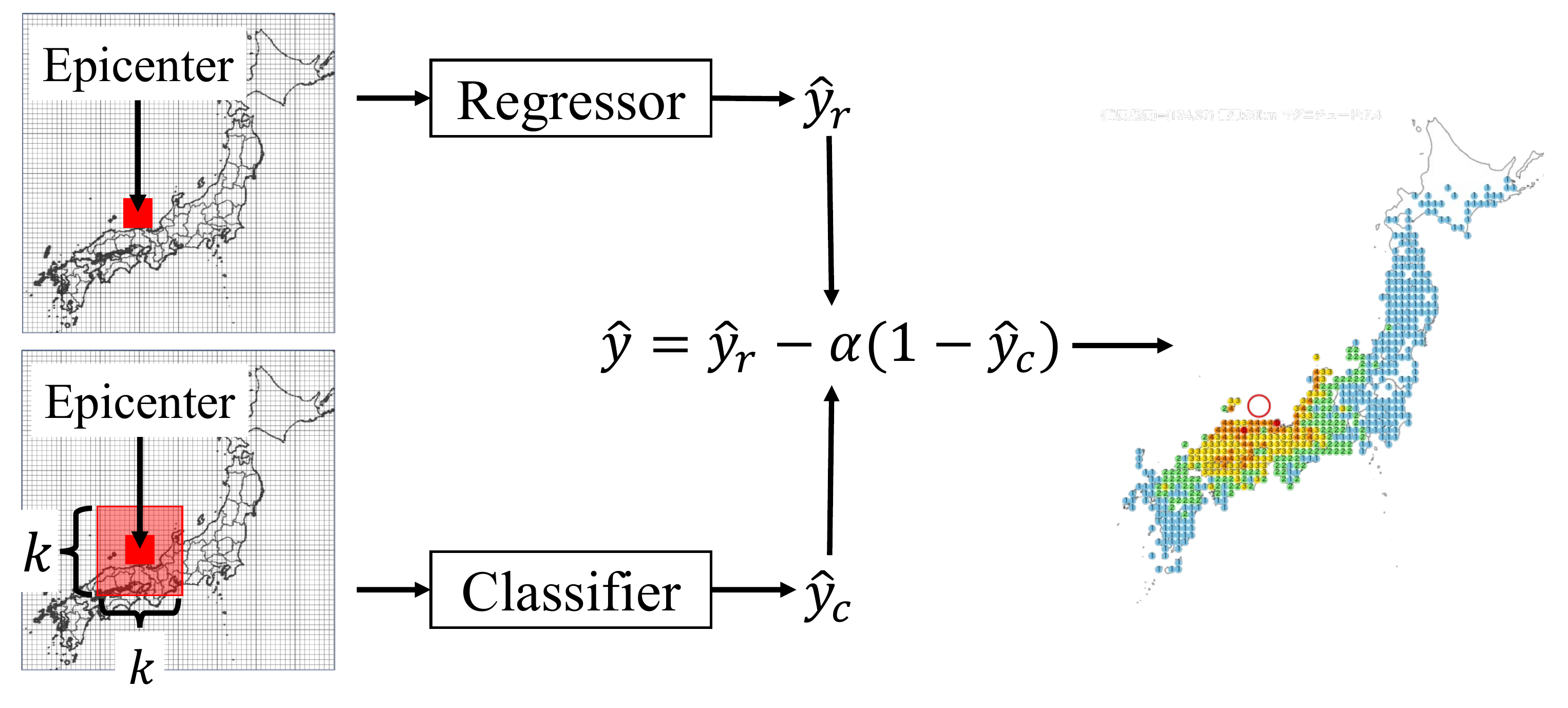}
        \caption{Input format and network schematic.}
        \label{fig:method}
        \vspace{-0.19cm}
    \end{figure}
    \paragraph{Dataset}We use 1,819 seismic intensity data of earthquakes of magnitude $5.0$ or greater between 1997 and 2019~\footnote{\url{https://www.data.jma.go.jp/eqev/data/bulletin/shindo_e.html}}. 
    The dataset was divided into 1,461 training data and 358 test data. The elements used in this study were the latitude, longitude, depth of the hypocenter and the earthquake's magnitude, seismic intensity on the Japan Meteorological Agency scale, and instrumental seismic intensity at each observation station.
    
    \paragraph{Data format}
    We used a Mercator map on a rectangular region of $30^\circ$N to $46^\circ$N, and $128^\circ$E to $146^\circ$E, divided into $64\times64$ square cells. The input and output data were also $64\times64$ 2-dimensional data, and each cell's seismic intensity was predicted.

    \paragraph{Method}We used regression and classification predictors in combination. The regressor predicts the instrumental seismic intensity at each cell~($\hat{y_\text{r}}$) . The classifier performs binary classification of whether a shock can be felt~($\hat{y_\text{c}}\in{\{0,1\}}$). Here, $\hat{y_\text{c}}=1$ for all the intensities greater than 0. We combine the regressor and classifier using the following equation:
    $\hat{y} = \hat{y_\text{r}} - \alpha \times (1-\hat{y_\text{c}})$,
    where $\hat{y}$ is the instrumental intensity predicted by the hybrid predictor and $\alpha$ is the scaling factor. This equation prevents overestimation of areas where shocks cannot be felt.
    
    \paragraph{Input data}
    Each input data is expanded to a $k\times k$ range centered on the epicenter for the classification model. Although the input data are not expanded in the regression forecasting, a large kernel size is set in the first convolution layer to simulate the expansion. In addition, the magnitude and depth of the input data were each multiplied by a power. This increases the dynamic range of the magnitude and depth values, and facilitates training. 
    The dataset did not contain seismic intensity data less than 0.5. Therefore, in the regression model, the data were ignored during the training. In contrast, in the classification model, such data were treated as having seismic intensities of 0.

\section{Results}
    \begin{figure*}[t]
        \begin{minipage}[b]{0.34\linewidth}
            \includegraphics[width = \linewidth]{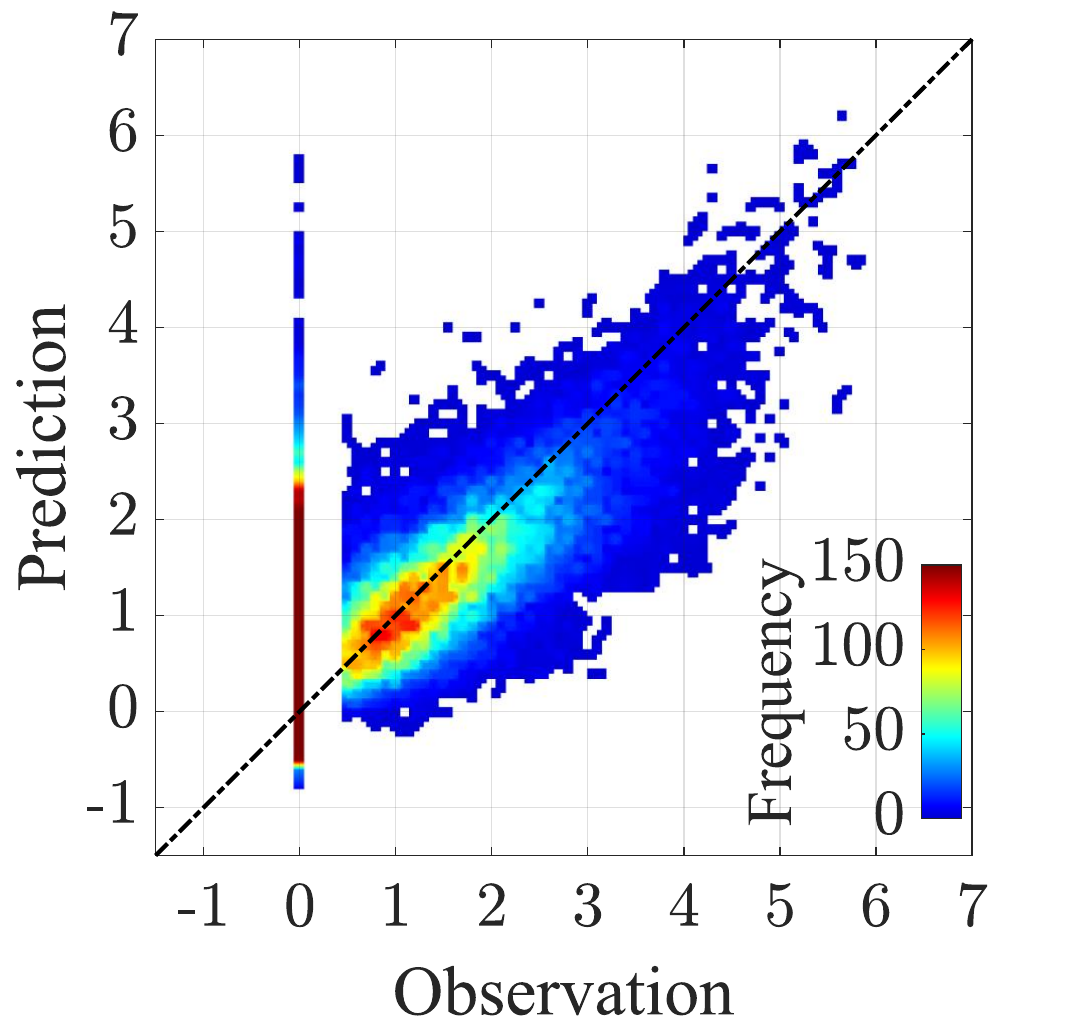}
            \caption{Relationship between observed and predicted intensity}
            \label{fig:pre-obs}
        \end{minipage}
        \hfill
        \begin{minipage}[b]{0.62\linewidth}
          \begin{minipage}[b]{0.49\linewidth}
            \centering
            \includegraphics[width = \linewidth]{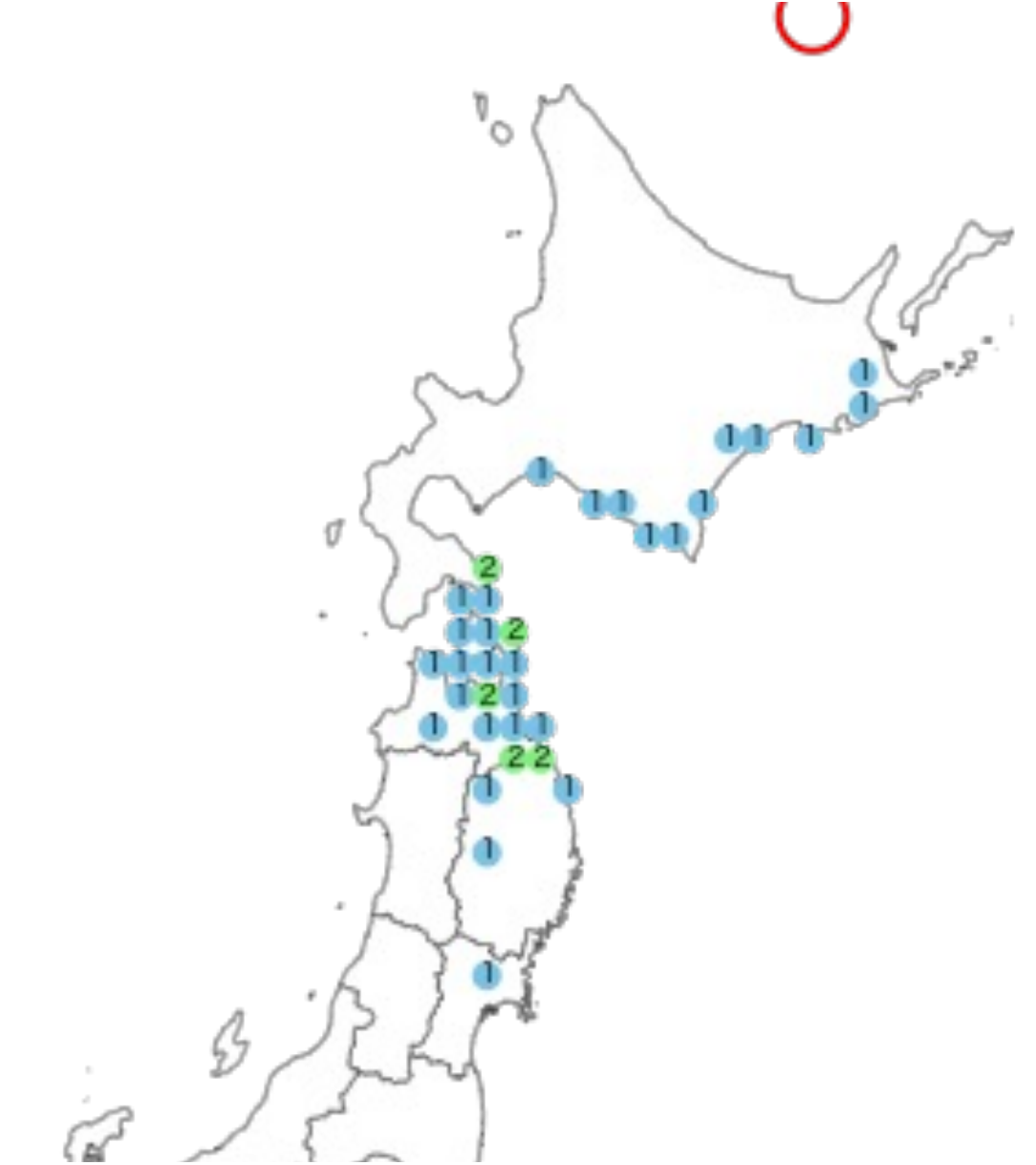}
            \subcaption{Observed}
          \end{minipage}
          \begin{minipage}[b]{0.49\linewidth}
            \centering
            \includegraphics[width = \linewidth]{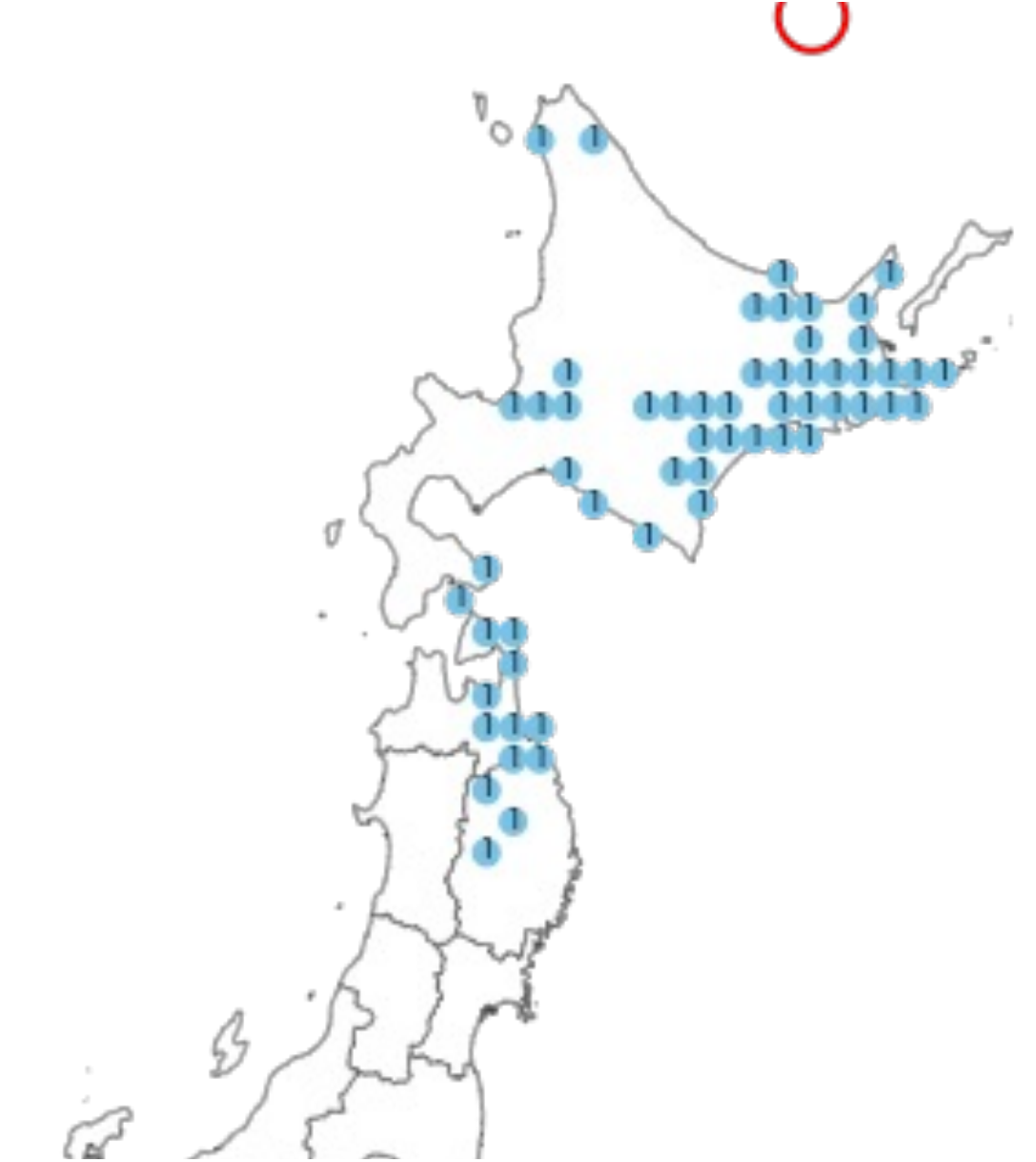}
            \subcaption{Predicted}
          \end{minipage}
          \caption{Abnormal distribution prediction. The hypocenter locates at $144^\circ$E (Longitude), $46^\circ$N (Latitude), and $387$km (Depth). The magnitude is $5.5$. }
          \label{fig:abnormal}
        \end{minipage}
    \end{figure*}
    \begin{table}[t]
        \caption{Mean squared error~(MSE), correlation coefficient~($r$), and F-score of each predictor.}
        \label{table:performance}
        \centering
        \begin{tabular}{cccc}
            \hline
            Predictor & MSE & $r$ & F-score \\
            \hline 
            \begin{tabular}{c}
                Regression
            \end{tabular}& $\mathbf{0.37}$ & $0.77$ & $0.44$\rule[0mm]{0mm}{0mm} \\
            \begin{tabular}{c}
                Classification
            \end{tabular}& $1.2$ & $0.59$ & $\mathbf{0.70}$\rule[0mm]{0mm}{0mm} \\
            \begin{tabular}{c}
                Regression + Classification
            \end{tabular}& $0.39$ & $\mathbf{0.78}$ &  $0.61$\rule[0mm]{0mm}{0mm} \\
            \hline
        \end{tabular}
        \vspace{-0.36cm}
    \end{table}
    In this section, we show that our method can accurately predict the seismic intensity and even the abnormal seismic intensity distributions. The architecture used was one fully connected layer for the classification model, one convolutional layer, and one fully connected layer for the regression model. Parameters $k=15$, kernel-size of the convolutional layer 125, and $\alpha =0.30$ were used. 
    As inputs for the classification predictor, the ninth-order terms of magnitude and the 1st-order term of depth were stored uniformly in the cells. For the regression predictor, the first- to fourteenth-order terms of magnitude and depth were stored. 
    
    Fig.~\ref{fig:pre-obs} shows the relationship between predicted and observed seismic intensities. There is a positive correlation, which indicates that our model can predict seismic intensity distributions.
    Predicted values are overall lower than the observed ones. This derives from the bias in the dataset.
    Tab.~\ref{table:performance} shows that our hybrid model recorded the best correlation coefficient~\footnote{Calculated using data from cells whose observed intensity is 0.5 or higher.} among the three models. 
    The proposed model did not achieve the best results in terms of MSE~\footnotemark[2] and F-score~\footnote{Calculated using data from cells at which observation stations exist.}; however, our model still achieved well-balanced values for MSE and F-score. The regression model had a low F-score, which indicates that it cannot accurately predict the extent of shaking. The classification model had a low correlation coefficient and a high F-score, indicating that it can accurately predict the extent of shaking, but not the intensity.
    
    
    Our method can also accurately predict abnormal seismic intensity distributions. An abnormal seismic intensity distribution for deep earthquakes is a phenomenon in which strong motions in the fore-arc are larger than those in the back-arc~\cite{iwakiri2011improvement}. The intensity at a distance from the epicenter may be stronger than that closer to the epicenter. Fig.~\ref{fig:abnormal} shows the distribution of observed abnormal distributions and their predicted results. The proposed method can predict the seismic intensity distribution, even in areas far from the epicenter.

    
\section{Conclusion}
    This study proposed a new method, developed by combining regression and classification models, for predicting seismic intensity distributions of earthquakes. The proposed method treats the seismic intensity distribution as 2-dimensional information.
    Our hybrid model achieved higher correlation coefficients and a well-balanced MSE and F-score compared with the only classification or regression models. Furthermore, we could accurately predict even abnormal seismic intensity distributions.
    In the future, we intend to develop a method that combines the model proposed in the present study with conventional physical models to further improve the accuracy of predicting seismic intensity distribution.

\bibliographystyle{IEEEtran}
\bibliography{main}

\end{document}